\documentclass[journal]{IEEEtran}
\usepackage[numbers,sort&compress]{natbib}
\usepackage[pdftex]{graphicx}
\graphicspath{{../pdf/}{../jpeg/}}
\DeclareGraphicsExtensions{.pdf,.jpeg,.png}
\usepackage{amsmath}
\usepackage{float}
\usepackage{longtable}
\usepackage{booktabs}
\usepackage{array}
\usepackage{makecell}
\usepackage{hyperref}
\usepackage{amsfonts}
\ifCLASSOPTIONcompsoc
  \usepackage[caption=false,font=normalsize,labelfont=sf,textfont=sf]{subfig}
\else
  \usepackage[caption=false,font=footnotesize]{subfig}
\fi

\begin{document}
%
% paper title
\title{Data-Driven Short-Term Voltage Stability Assessment Based on Spatial-Temporal Graph Convolutional Network}
% author names and IEEE memberships
\author{Yonghong~Luo,
        Chao~Lu,~\IEEEmembership{Senior~Member,~IEEE,}
        Lipeng~Zhu,
        and~Jie~Song% <-this % stops a space
\thanks{Chao Lu is with the Department
of Electrical Engineering, Tsinghua University, China,
e-mail: luchao@tsinghua.edu.cn. This work was supported
by National Natural Science Foundation of China (U1766214, 51677097).}% <-this % stops a space
}
% The paper headers
%\markboth{IEEE transactions on}%
%{Shell \MakeLowercase{\textit{et al.}}: Bare Demo of IEEEtran.cls for IEEE Journals}
\maketitle
% As a general rule, do not put math, special symbols or citations
% in the abstract or keywords.
\begin{abstract}
Post-fault dynamics of short-term voltage stability
(SVS) present spatial-temporal characteristics, but the existing
data-driven methods for online SVS assessment fail to incorporate such characteristics into their models effectively. Confronted
with this dilemma, this paper develops a novel spatial-temporal
graph convolutional network (STGCN) to address this problem. The proposed STGCN utilizes graph
convolution to integrate network topology information into the
learning model to exploit spatial information. Then, it adopts one-dimensional convolution to exploit temporal information. In
this way, it models the spatial-temporal characteristics of SVS
with complete convolutional structures. After that, a node layer
and a system layer are strategically designed in the STGCN
for SVS assessment. The proposed STGCN incorporates the
characteristics of SVS into the data-driven classification model. It can result in higher assessment accuracy, better robustness and
adaptability than conventional methods. Besides, parameters
in the system layer can provide valuable information about
the influences of individual buses on SVS. Test results on the
real-world Guangdong Power Grid in South China verify the
effectiveness of the proposed network.
\end{abstract}

% Note that keywords are not normally used for peerreview papers.

\IEEEpeerreviewmaketitle
\section{Introduction}
\subsection{Reseach background}
\IEEEPARstart{W}{ITH} the rapid growth of power consumption, large-scale integration of renewable energy sources, and the increasing penetration of dynamic loads, short-term voltage instability has become the prominent problem in power systems \cite{VS}. When suffering
from large disturbances, power systems may experience short-term voltage instability and trigger blackouts \cite{non:shellCTANpage}. Blackouts can bring about huge economic losses and social impacts. Therefore, it is crucial to correctly assess voltage stability to take measures timely and prevent the occurrence of blackouts.

Recently, the rapid development of the wide-area measurement system (WAMS) has greatly promoted situational awareness of
power systems. Phasor measurement
units (PMUs) are widely deployed in the high-voltage stations of power systems. The PMU data is synchronized by the Global Position System (GPS) and can provide the real-time states of power systems \cite{PMU}. The massive PMU data makes it feasible to conduct PMU data-driven
short-term voltage stability (SVS) assessment, and it creates opportunities to provide SVS assessment results in a short time.
\subsection{Literature review}
In general, there are mainly three categories of approaches for PMU data-driven stability assessment over past decades, namely the practical criteria, the stability mechanism-based methods, and machine learning-based methods.
The widely accepted practical criterion is: when the duration of any bus voltage under a threshold exceeds the preset time \cite{WECC}, \cite{china}, the system is assessed as unstable. However, the threshold and the preset time all depend on the operating experiences of dispatchers, without any support from theory or enough data. Therefore, practical criteria lack reliability and adaptation. One of the typical stability mechanism-based methods is the maximum Lyapunov exponent (MLE) method \cite{non:dasgupta2013real-time}. This method
can conduct SVS assessment with the sign of the MLE value.
However, as the MLE value often oscillates around 0, it cannot
provide a reliable result. In \cite{2017IM}, the recovery time of equivalent induction motor rotation speed is estimated to assess SVS status. The accuracy of this method highly depends on the identification of induction motor parameters, while the identification of induction motor parameters is also a very difficult problem.

Machine learning-based methods have received much attention among researchers due to their superior performances. These methods learn the mapping between the input data and the final assessment results through the training dataset. Then, given the practical input data, the well-trained model can provide the assessment results based on the learned mapping. According to the use of input information, the existing machine learning-based assessment methods can be divided into three categories. The methods of the first category take single snapshots as the learning inputs, such as artificial neural network \cite{ANN}, ensemble model of neural networks with random weights (NNRW) \cite{DBLP:journals/tnn/XuZZDWYW16}, and ensemble model of extreme learning machine (ELM)
\cite{DBLP:journals/tii/ZhangXDZ19}. However, with only single snapshots of time series, the
temporal dynamics of SVS cannot be
well characterized. To exploit the time-varying characteristics, the methods of the second category take post-fault
time series as the learning input for stability assessment, such as shapelet-based 
methods \cite{zhu2016time}, \cite{zhu2017imbalance}, long short-term memory (LSTM) model \cite{yu2018intelligent}, and random vector functional link (RVFL) \cite{zhang2019a}. Nevertheless, apart from the time-varying characteristics, short-term voltage instability also presents spatial distribution characteristics \cite{zhu2020spatialtemporal}. Therefore, the methods of the third category take spatial-temporal
information for stability assessment. The spatial-temporal characteristics of SVS have been neglected for years by most machine learning-based methods. To the best of the
authors’ knowledge, there is only one attempt that incorporates
spatial-temporal information, namely
the spatial-temporal shapelet learning method \cite{zhu2020spatialtemporal}. This method
incorporates spatial information into the learning model based
on the geographical location information of the buses. However, the geographical locations may not exactly reflect the electrical distance
between the buses, and the model
based on geographical locations may not be accurate enough.

\subsection{Motivation and contribution}
In fact, short-term voltage instability presents salient spatial-temporal characteristics. The affected region of short-term voltage instability exhibits spatial distribution characteristics and the low-voltage region reveals locality over topology. This phenomenon derives from the fact that reactive power cannot be
transmitted over long distances and low voltage buses mainly
affect electrically neighboring buses. Besides, the affected region of short-term voltage instability changes over time. However, these spatial-temporal characteristics haven't been effectively incorporated into the learning model by the existing
machine learning-based SVS assessment methods.

The key of getting an assessment model with good performances lies in the determination of input data and the design of the learning model. However, in the input data aspect, most of the existing methods fail to fully utilize the spatial-temporal information of the post-fault dynamics in SVS. Logically, the spatial-temporal characteristics in the post-fault dynamics haven't been incorporated into the learning model. Once the topology information and temporal dynamics of post-fault measurements are combined together,
the spatial-temporal characteristics of SVS could be fully exploited,
resulting in a more credible SVS assessment model.

Therefore, in this paper, spatial-temporal
graph convolutional network (STGCN) is developed to fully exploit the spatial-temporal characteristics of SVS and improve the performances of the SVS assessment model. Graph neural network is utilized to integrate
topology information into the learning model and exploit the spatial information in SVS. Graph neural network
has been widely used in social networks, transportation, chemistry, fault location in power
distribution systems \cite{chen2020fault}, etc. It has shown great advantages in tackling data residing on graphs \cite{battaglia2018relational}, \cite{wu2019a}. The
foundation of graph neural network is graph convolutional
network (GCN), which is powerful for incorporating spatial
information. Therefore, it is adopted as the graph convolutional layer of the proposed STGCN. As for spatial-temporal information
incorporation, there are mainly four ways by exploiting graph
convolution: 1) adding a one-dimensional convolutional layer
behind the graph convolutional layer \cite{yu2018spatiotemporal}; 2) adding a long
short-term (LSTM) layer or gated recurrent unit (GRU) behind
the graph convolutional layer \cite{yao2018deep}; 3) modifying the original
LSTM or GRU, which is to replace the fully connected layer
in LSTM or GRU by graph convolution \cite{seo2016structured}; 4) representing
temporal correlations as new edges of the graph and constructing a new graph with spatial-temporal correlations \cite{yan2018spatial}.
The second and third methods can achieve the incorporation of 
spatial-temporal information. However, the training time with LSTM or
GRU-based method is longer than that with a one-dimensional
convolutional layer. The fourth method constructs graph to
treat temporal correlation with spatial correlation equally.
It may not make sense due to the distinction between
temporal correlations and spatial correlations.

Therefore, the one-dimensional convolutional layer is further adopted to enable
spatial-temporal feature extraction from the hidden states of
the graph convolutional layer. In this way, it incorporates
spatial-temporal information with complete convolutional
structures. Then, with consideration of SVS characteristics,
a node layer is strategically designed to generate node representations for the buses. Based on the
node representations, the system layer employs the simplified
differentiable pooling \cite{diffpool} to provide the assessment
result.

The main contributions of this paper are:
\begin{itemize}
\item A model framework with spatial-temporal information incorporation is developed in this paper. It successfully bridges the temporal data and topology information together to construct a model with higher performance.

\item The spatial-temporal characteristics of SVS are first modeled with graph convolution and one-dimensional convolution. Combined with SVS characteristics, the proposed model can achieve higher accuracy, better robustness and adaptability.

\item The influences of individual buses on SVS are provided and analyzed with the parameters in the designed system layer.

\item Most existing works use the New England 39-bus system as the test system, which may not be large enough to fully test the performance of the learning model. In this paper, we use the real-world Guangdong Power Grid as the test system, which takes 101 high-voltage buses into account.
\end{itemize}

The remainder of this paper is organized as follows. A brief
introduction of the proposed network for SVS assessment is
given in Section \uppercase\expandafter{\romannumeral2}. In Section \uppercase\expandafter{\romannumeral3},  the design of STGCN is
illustrated in detail. The real-world Guangdong Power Grid is
utilized to test the performances of the proposed network in
Section \uppercase\expandafter{\romannumeral4}. Finally, conclusions are presented in Section \uppercase\expandafter{\romannumeral5}. 

\section{Framework of STGCN for SVS assessment}
\subsection{Online SVS assessment problem}
SVS assessment is a tricky problem that has received long-term attention among researchers. Due to the high dimensionality, time-varying characteristics, and strong nonlinearity of
SVS, the post-fault dynamics of SVS are very complex. Therefore, it
is hard to achieve accurate online SVS assessment.

In fact, in large-scale complex power systems, the characteristics of short-term voltage instability can not only be
reflected in post-fault responsive trajectories, but also spatial
distribution. When severe faults happen, the affected regions
of short-term voltage instability reveal obvious spatial distribution characteristics over topology, as shown in Fig. 1. This illustrative voltage
distribution map is based on a snapshot of the post-fault voltage magnitude
time series in the IEEE 39-bus test system. The legend on
the right side of the map illustrates the correspondence
between colors and voltage values. The region where the power system is located can be considered as a 2-D geographical space, and the map can be depicted with spatial voltage interpolation. As for the implementation of spatial voltage interpolation, the locations with real buses have their corresponding voltage values in the geographical space at the beginning. Spatial voltage interpolation means that the locations without real buses can be filled with virtual voltage values, and these virtual voltage values are estimated by geographical distance weighted interpolation. The detailed information about the spatial voltage interpolation can be found in \cite{weber2000voltage}. 

As shown in Fig. 1,
SVS exhibits salient spatial distribution characteristics over
topology, and the low-voltage region reveals locality. If these characteristics are
considered in the design of the learning model, it is feasible
to achieve SVS assessment with higher performances.

\begin{figure}[h]
\centering{\includegraphics[height=6.8cm,width=8.5cm]{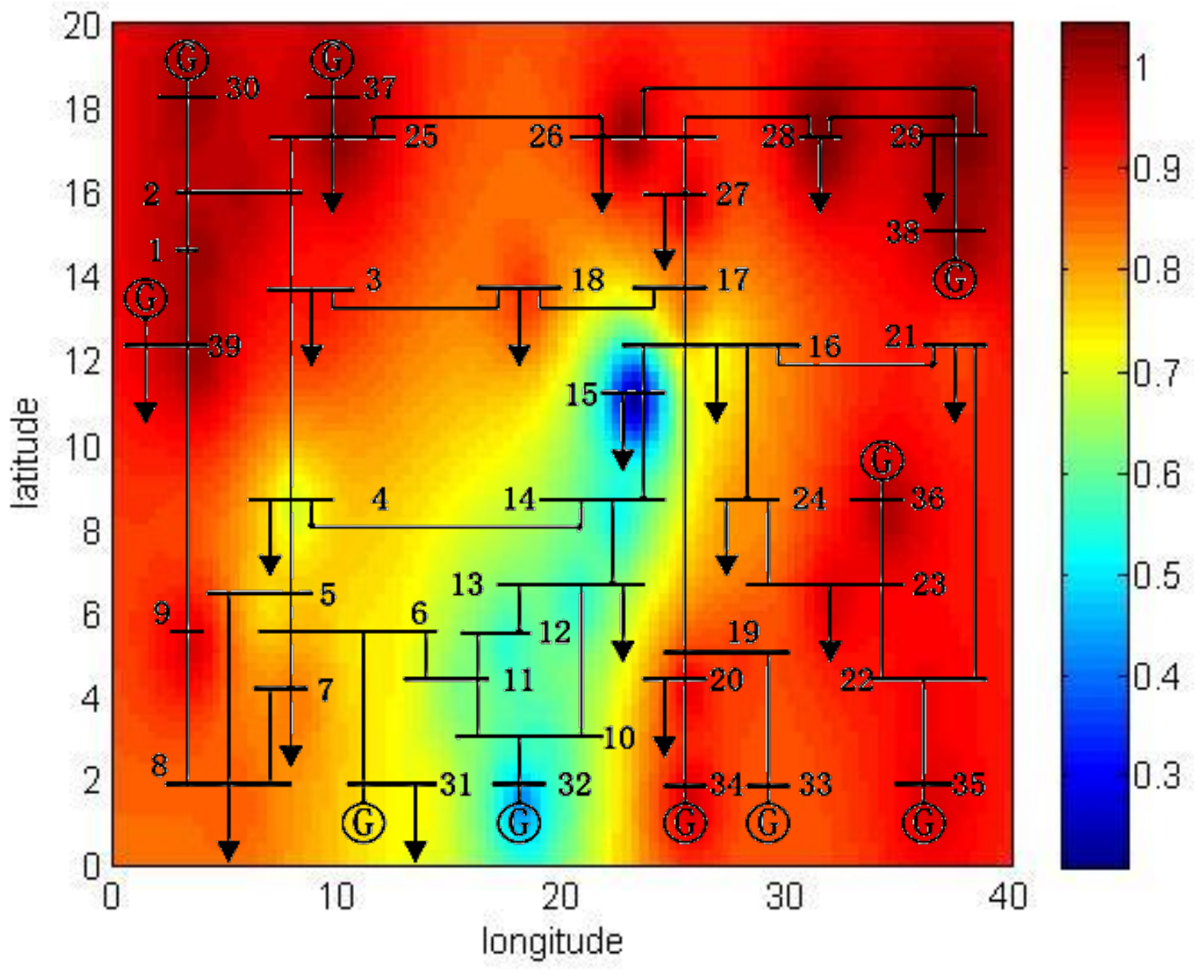}}
\caption{Illustrative voltage distribution map
\label{fig_sim}}
\end{figure}

\subsection{Graph neural network}
Graph neural networks refer to a class of neural networks
that can incorporate topology into the learning models and
take advantage of graph dataset characteristics. They impose
strong relational inductive biases on the network structure
and are more appropriate for graph datasets than traditional
convolution networks. Relational inductive biases refer to
the constraints on the relationships and interactions among
variables in a learning process. They can also be regarded
as the assumptions about the data generation process or
the space of solution \cite{battaglia2018relational}. Proper relational inductive biases in graph neural networks can improve model accuracy and
generalization ability.

GCNs are the foundation of various graph neural networks. They can incorporate spatial information by graph convolution. GCNs fall into two categories of approaches,
namely spectral theory-based approaches \cite{bruna2013spectral,defferrard2016convolutional,kipf2016semi-supervised} and spatial-based approaches \cite{scarselli2009the,gilmer2017neural,hamilton2017inductive}. Spectral theory-based graph convolution is
developed from graph signal processing domain. Its basic idea
is to realize graph convolution by introducing filters from the
perspective of graph signal processing. Spatial-based graph
convolution is similar to the traditional convolution network. It is implemented as aggregating features from neighbors. It
is more flexible, but no theoretical basis. As spectral theory-based graph convolution has a solid theoretical basis, it is
adopted to incorporate spatial information into the learning
model.

Spectral theory-based graph convolution networks study the
properties of a graph by eigenvalues and eigenvectors of
the normalized Laplacian matrix. The normalized Laplacian
matrix is the mathematical representation of a graph, defined
as
\begin{align}
L=I_n-D^{-1/2}WD^{-1/2}
\end{align}
where $W$ is the adjacency matrix with weights, representing the
connection relationships among nodes in the graph, $D$ is a diagonal
matrix of the corresponding nodal degrees $D_{i, i}=\sum_{j}\left(W_{i, j}\right)$, and $I_n$ is an identity matrix. The normalized Laplacian matrix
has real symmetric semi-definite properties, which can be
factored as 
\begin{align}
L=U\Delta U^T
\end{align}
where $\Delta$ is a diagonal matrix composed of eigenvalues and $U$ is the matrix of eigenvectors ordered by eigenvalues. The graph
Fourier transform to a signal x is defined as $F(x) =U^Tx$. $U^T$ is regarded as the graph Fourier basis. 

The graph convolution of the input signal $x$ with filter $g$ is defined as 
\begin{align}
x*Gg=U(U^Tx\bigodot U^Tg)
\end{align}
where $\bigodot$ denotes the Hadamard product. If the filter is denoted
as $g_{\theta}=\operatorname{diag}\left(U^{T} g\right)$, the graph convolution can be simplified as

\begin{align}
x * G g=U g_{\theta} U^{T} x
\end{align}
All graph convolution networks based on spectral theory follow this definition, but the filters $g_{\theta}$ are different. 

\subsection{Novel framework for SVS assessment}
To exploit the spatial distribution characteristics of SVS,
graph convolution is utilized to integrate topology information
into the learning model. Then, the spatial-temporal information
incorporation block with the graph convolutional layer is
designed to extract the spatial-temporal features of SVS. In
fact, the stacking number of the spatial-temporal information
incorporation blocks can represent the reception fields of the buses. In order to capture spatial-temporal features from different reception fields, several spatial-temporal information incorporation blocks are stacked, and the fusion
information from each spatial-temporal information incorporation block is utilized for SVS assessment.

As a matter of fact, voltage stability is regarded as load
stability for a long time. Although it may not be accurate
enough, the state of load buses can affect voltage stability to
a large extent. Therefore, the node layer block is designed to
generate the node representations of the buses after the spatial-temporal information incorporation
blocks. Then, based on the node representations, the system
layer block is utilized to provide the final SVS assessment
result.

The framework of the proposed STGCN to conduct the
SVS assessment task is shown in Fig. 2. The input data will
be processed by three parts, the spatial-temporal information
incorporation blocks, the node layer block, and the system
layer block. This framework incorporates the characteristics
of SVS into the learning model and is more appropriate for
online SVS assessment. The detailed information about these blocks will be illustrated in Section \uppercase\expandafter{\romannumeral3}.

\begin{figure}[h]
\centering{\includegraphics[height=4.6cm,width=8.8cm]{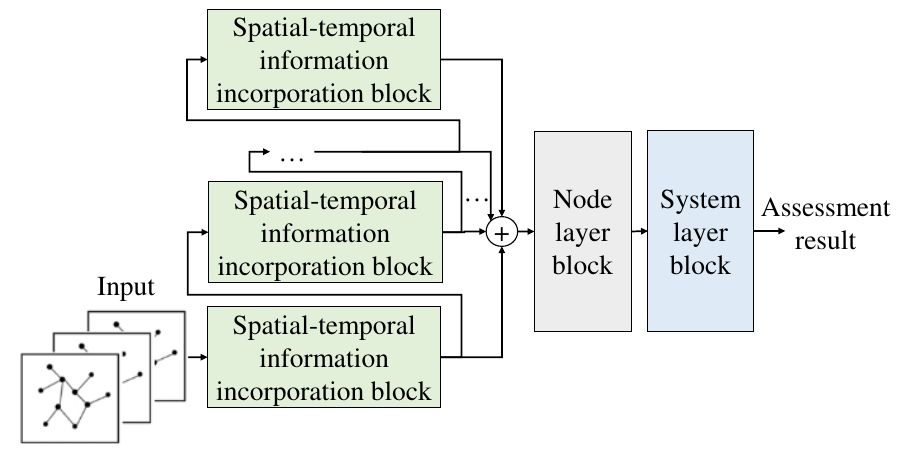}}
\caption{The framework of STGCN
\label{fig_sim}}
\end{figure}

The implementation of the proposed network for SVS assessment can be divided into two stages, the offline training stage and the online assessment stage. At the offline training stage, a large number of samples are generated with electromechanical transient analysis programs. Then, the database can be formed with post-fault time series and topology of these samples as the inputs and the corresponding stability status as the output. Time series for the inputs consist of
voltage magnitude, active power injection, and reactive power
injection time series. With the prepared database, STGCN can
be trained with gradient descent algorithms, such as the Adam
algorithm \cite{Adam}. When the testing accuracy of STGCN reaches a preset threshold, the trained STGCN will be saved for online assessment. At the online assessment stage, post-fault PMU measurements and topology of the observed region can be obtained as the inputs. Then, the trained STGCN is utilized to conduct SVS assessment. If the assessment result shows that the system is stable, then the model continues to monitor. If the assessment result shows that the system is unstable, warning signals will be sent out to remind operators to take measures. 

\section{Design of STGCN}
\subsection{Definition of the input layer}
SVS assessment is a typical multivariate time series classification task. In this paper, it is assumed that voltage magnitude
time series $V_t$, active power injection time series $P_t$ and
reactive power injection time series $Q_t$
are available and of great importance for SVS. Therefore,
these multivariate time series are utilized as the inputs of the
proposed network
\begin{align}
V_t & = \begin{bmatrix} v_1 & v_2 & ... & v_n \end{bmatrix} \\
P_t & = \begin{bmatrix} p_1 & p_2 & ... & p_n \end{bmatrix} \\
Q_t & = \begin{bmatrix} q_1 & q_2 & ... & q_n \end{bmatrix} 
\end{align}
where $t=1,2,...,N$, $N$ is the number of time points for stability assessment, $n$ is the number of observed buses.

As short-term voltage instability presents spatial distribution
characteristics, the topology $W$ of the observed region is also
extracted as the input. More specifically, the topology matrix
$W$ is composed of a node admittance matrix. When the
topology of the observed region changes, the topology matrix
also changes, thereby adapting to the changes of topology. 
\begin{align}
W & = \begin{bmatrix}w_{11} & w_{12} & ... & w_{1n} \\
 w_{21} & w_{22} & ... & w_{2n} \\
 ... & ... & ... & ... \\
 w_{n1} & w_{n2} & ... & w_{nn} \\
\end{bmatrix} 
\end{align}

The illustrative figure for the input data of STGCN is shown in Fig. 3.

\begin{figure}[h]
\centering{\includegraphics[height=4.1cm,width=6.2cm]{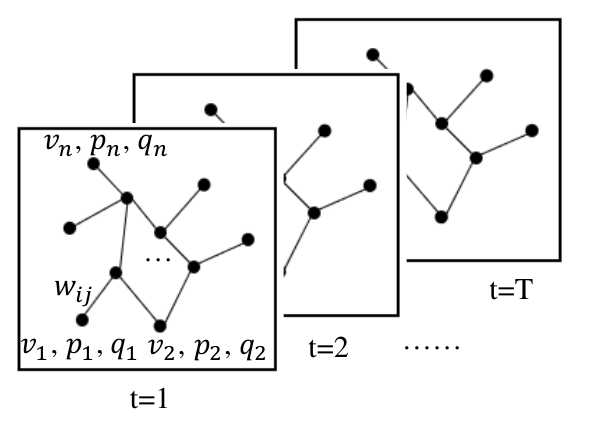}}
\caption{The input data for ST-GCN
\label{fig_sim}}
\end{figure}

\subsection{Spatial-temporal information incorporation blocks}
Spatial-temporal information incorporation blocks are
formed by stacking several spatial-temporal information incorporation blocks and utilizing the fusion information from
each spatial-temporal information incorporation block.

Each spatial-temporal information incorporation block consists of four parts, a graph convolutional layer, a temporal
convolutional layer, a normalization layer, and a dropout
layer. The graph convolutional layer is adopted to incorporate spatial information and extract
spatial features. The temporal convolutional layer is adopted
to incorporate temporal information and extract temporal features. In order to prevent over-fitting,
a normalization layer and a dropout layer are added to improve the performances of the proposed network. The detailed illustration of one spatial-temporal information incorporation
block is shown in Fig. 4.

\subsubsection{The graph convolution layer}
As spectral theory-based
graph convolution has a solid theoretical basis, it is adopted
to incorporate spatial information. The first spectral convolution neural
network is proposed by Bruna \cite{bruna2013spectral}. The corresponding filter is defined
as $g_{\theta}=\theta_{i,j}^k$. As the computation of the graph Fourier basis
can cost lots of time, Defferrard proposes Chebyshev spectral
convolution neural network (ChebNet) \cite{defferrard2016convolutional}. This method defines filters as Chebyshev polynomials of the eigenvalue diagonal matrix. When only using the first polynomial, it turns
into the first-order approximation of ChebNet \cite{kipf2016semi-supervised}. Due to
the less computational complexity and locality of Chebyshev
filter, ChebNet is adopted to extract the spatial features in SVS
assessment.

ChebNet defines filters as Chebyshev polynomials of the
eigenvalue diagonal matrix
\begin{align}
g_{\theta}={\sum}_{i=0}^K{\theta}_iT_i(\tilde{\Delta})
\end{align}
where $\tilde{\Delta}=2\Delta/{\lambda}_{max}-I_n, T_i(\tilde{\Delta})=2\tilde{\Delta}T_{i-1}(\tilde{\Delta})-T_{i-2}(\tilde{\Delta})$ with $T_0(\tilde{\Delta})=1$, $T_1(\tilde{\Delta})=\tilde{\Delta}$. ${\lambda}_{max}$ is the  maximal eigenvalue, ${\theta}_i$ is the learnable parameter and $K$ is the Chebyshev order.

The input signal for graph convolution can be voltage magnitude time series, active power injection time series and reactive power injection time series. The graph convolution of voltage magnitude $V_t$ with filter $g_{\theta}$ in a time step can be represented as
\begin{align}
V_t*Gg_{\theta}&=U({\sum}_{i=1}^K{\theta}_{Vi}T_i(\tilde{\Delta}))U^TV_t \\
&={\sum}_{i=0}^K{\theta}_{Vi}T_i(\tilde{L})V_t
\end{align}
The graph convolution of voltage magnitude $P_t$ with filter $g_{\theta}$ in a time step can be represented as
\begin{align}
P_t*Gg_{\theta}&=U({\sum}_{i=1}^K{\theta}_{Pi}T_i(\tilde{\Delta}))U^TP_t \\
&={\sum}_{i=0}^K{\theta}_{Pi}T_i(\tilde{L})P_t
\end{align}
The graph convolution of voltage magnitude $Q_t$ with filter $g_{\theta}$ in a time step can be represented as
\begin{align}
Q_t*Gg_{\theta}&=U({\sum}_{i=1}^K{\theta}_{Qi}T_i(\tilde{\Delta}))U^TQ_t \\
&={\sum}_{i=0}^K{\theta}_{Qi}T_i(\tilde{L})Q_t
\end{align}
where ${\theta}_{V}$, ${\theta}_{P}$, ${\theta}_{Q}$ are the learned paprameters in the graph convlutional layer, $T_{i}(\tilde{L})=2 \tilde{L} T_{i-1}(\tilde{L})-T_{i-2}(\tilde{L})$ with $T_{0}(\tilde{L})=1$, $T_{1}(\tilde{L})=\tilde{L}$, $\tilde{L}=2 L / \lambda_{\max }-I_{n}$, $L=I_n-D^{-1/2} WD^{-1/2}$ and $W$ indicates the topology of the observed region.

$T_{i}(\tilde{L})$ is Chebyshev filter matrix, which is the key element
to realize graph convolution. In the graph convolutional layer,
each bus updates its node information according to the information in its reception field. The locality of Chebyshev filter
matrix ensures that the buses in the reception field of one bus
are its neighboring buses. It is consistent with the spatial
distribution characteristics of SVS.

\begin{figure*}[h]
\centering{\includegraphics[height=5.8cm,width=18.2cm]{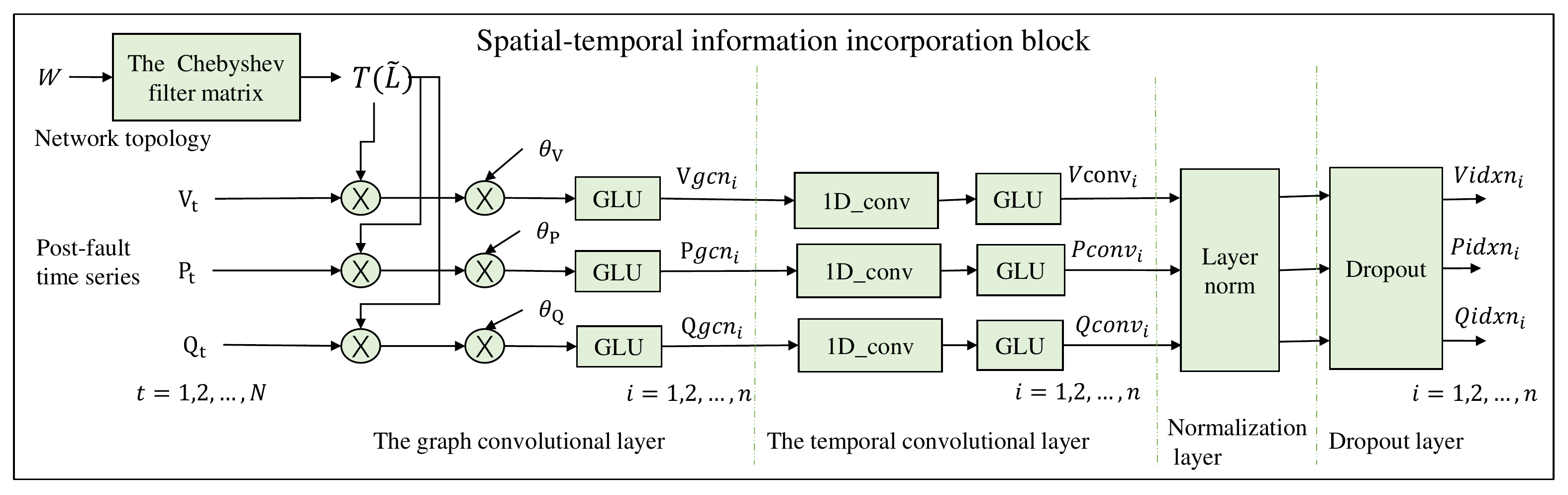}}
\caption{The spatial-temporal information incorporation block
\label{fig_sim}}
\end{figure*}

After the graph convolution of the input time series, the
activation function, such as the gated linear unit (GLU) \cite{glu}, is utilized to model the nonlinear characteristics of SVS.
$Vgcn_i$, $Pgcn_i$, $Qgcn_i$ are the data processed by the graph convolutional layer, $Vgcn_i = [vgcn_1\quad vgcn_2\quad ...\quad vgcn_N]^T$,
$Pgcn_i = [pgcn_1\quad pgcn_2\quad ...\quad  pgcn_N]^T$, $Qgcn_i = [qgcn_1\quad qgcn_2\quad ...\quad qgcn_N]^T$.

\subsubsection{The temporal convolution layer}
After the extraction
of spatial features, the one-dimensional convolution is adopted
to incorporate temporal information and extract temporal features. After that, the activation function,
such as gated linear unit (GLU), is utilized to model the
nonlinear characteristics of SVS
\begin{align}
&Vconv_i=GLU(1D\_conv(Vgcn_i)) \\
&Pconv_i=GLU(1D\_conv(Pgcn_i)) \\
&Qconv_i=GLU(1D\_conv(Qgcn_i)) 
\end{align}
where ${Vconv}_i$,${Pconv}_i$,${Qconv}_i$ are the data processed by the temporal convolution layer, $i=1,2,..,n$. 

Then, the normalization layer and dropout layer are added to prevent over-fitting, and the data are transformed as $Vidxn_i$, $Pidxn_i$, $Qidxn_i$.

When stacking several spatial-temporal information incorporation blocks, to distinguish the data processed by different
blocks, the data $Vidxn_i$, $Pidxn_i$, $Qidxn_i$ are further denoted
as $Vidxn_{i,j}$, $Pidxn_{i,j}$, $Qidxn_{i,j}$, and $j$ is utilized for distinguishing different blocks.

\subsubsection{Spatial-temporal convolution incorporation blocks}
To capture spatial-temporal features from different reception
fields, $L_c$ spatial-temporal information incorporation blocks
are stacked, and the fusion information from each spatial-temporal information incorporation block is utilized for SVS
assessment.
\begin{align}
&Pidxz_i=\sum_{j=1}^{L_c}Pidxn_{i,j} \\
&Qidxz_i=\sum_{j=1}^{L_c}Qidxn_{i,j} \\
&Vidxz_i=\sum_{j=1}^{L_c}Vidxn_{i,j} 
\end{align}
where $i=1,2,..,n$, ${Vidxz}_i$, ${Pidxz}_i$, ${Qidxz}_i$ are the data processed by the fusion of the spatial-temporal information incorporation blocks.
\subsection{Node layer block and system layer block}
\subsubsection{Node layer block}
After extracting the spatial-temporal
features of different channels, namely voltage magnitude
channel, active power injection channel, and reactive power
injection channel, new representations of different channels
on each bus can be obtained.

The node layer block is utilized to convert them into a single
representation on each bus by applying a weighted summation
on different channels. Then, the data are processed by a normalization layer and
the absolute values are taken as their node representations
\begin{align}
Snode_i=|Layernorm(&{\psi}_1*Pidxz_i+{\psi}_2*Qidxz_i \notag \\
&+{\psi}_3*Vidxz_i)|
\end{align}
where ${Snode}_i$ is the node representation of bus $i$, ${\psi}_1$, ${\psi}_2$, ${\psi}_3$ are the weights in the node layer representation block. 

\subsubsection{System layer block}
Based on the node representations,
the SVS status of the target region can be obtained in
three ways, namely max/mean pooling, multi-layer perceptron
plus softmax function, and differentiable pooling layer \cite{diffpool}.
The differentiable pooling layer can generate hierarchical
representations of a graph, showing better performances than
other graph pooling methods. Based on the node representation, the differentiable
pooling layer is simplified as the system layer to provide the
SVS assessment result.
It is defined as follows
\begin{align}
&Ssys=softmax(SNODE*softmax(Sb)^T)  
\end{align}
where $SNODE=[Snode_1\quad Snode_2\quad ...\quad Snode_n]$, $Sb$ is the dense learned assignment matrix, $Sb\in\mathbb{R}^{2\times n}$, $Ssys$ is the SVS assessment result, $Ssys\in\mathbb{R}^{1\times 2}$. Softmax function is used as the final function of the network. It can provide probability values for different categories \cite{softmax}, \cite{softmax1}. The predicted category of a sample is the corresponding category of the column with the highest probability value. For binary classification in SVS assessment, the probability value in the first column of $Ssys$ is set to represent the probability of stable status, and the probability value in the second column of $Ssys$ is set to represent the probability of unstable status. Therefore, if the element of
$Ssys$ in the first column is greater than the element of $Ssys$ in the
second column, the system is stable, otherwise, it is unstable.
Cross entropy loss function is adopted as the loss function of
the proposed network.

The elements of $softmax(Sb)^T$ in the first column
minus the elements of $softmax(Sb)^T$ in the second
column are regarded as the learned parameters $S$ in the system
layer. With the node representations $SNODE$ and learned parameters $S$ in the system layer, the SVS status can also be obtained by multiplying these two matrices.
When the final result is positive, the
target region is stable, otherwise, it is unstable. As the
absolute values are taken as the node representations, the sign
of parameters $S$ in the system layer can directly affect the
assessment result. Therefore, the positive/negative weight
values $S$ can indicate the beneficial/detrimental influences of
corresponding buses on SVS.

\section{Case study}
\subsection{System description and simulation setting}
The proposed network is tested on the real-world Guangdong Power Grid. Guangdong Power Grid is a typical
receiving-end system of China Southern Power Grid. It is prone to suffer from the voltage instability problem. Therefore, Guangdong Power Grid is utilized for SVS assessment. The backbone
structure of Guangdong Power Grid is shown in Fig. 5. There
are 101 high-voltage buses in Guangdong Power Grid. All
high-voltage buses in Guangdong Power Grid are assumed to
be installed with PMUs. These PMU measurements are utilized for online monitoring.
All loads in the system are represented by the composite load
model consisting of induction motor and static loads.

Samples are generated by PSD-BPA software, which is the
power system simulation software widely used by Chinese
power companies. 5400 cases are generated with PSD-BPA
software by setting different fault locations, different fault
clearing time, and different induction motor ratios. More
specifically, the fault clearing time is set to 0.1, 0.2, 0.3, 0.4,
0.5 seconds. The induction motor ratio is set to 0.3, 0.5, 0.7,
0.9. Three-phase short-circuit faults are imposed on 270
different fault locations which are randomly selected from
the high-voltage transmission lines in Guangdong Power Grid.
Besides, to consider topology changes, 600 cases are generated
with PSD-BPA software by setting different topology changes,
different fault clearing time, and different induction motor
ratios. More specifically, the fault clearing time is set to
0.5, 1.0, 1.5 seconds to simulate severe situations, and the
induction motor ratio is set to 0.7 and 0.9. Each of these
topology changes in the dataset contains a one-part change
of the original topology. In addition, the transient simulation
in each case lasts 10 seconds after fault clearance. These 6000
cases are randomly mixed and utilized to test the performances
of the proposed network. Besides, more cases are generated in the later subsections to fully test the performances of the proposed model under different kinds of environments.

For each case, the 10-second post-fault trajectory data is
used to provide an output label according to the stability status.
The topology, namely the node admittance matrix, is extracted
to form the input topology matrix. The data of one second after
fault clearance, including voltage magnitude, active power
injection, and reactive power injection time series, are used
as the input time series. In fact, the length of the observation
window for SVS assessment can affect classification accuracy.
The longer the observation window, the higher the assessment
accuracy. However, SVS assessment in practice also requires
earliness. The observation window used in this article is a typical setting in the related researches \cite{zhu2020spatialtemporal}. Besides, the proposed method is also available for temporal-adaptive implementation referring to the scheme proposed in \cite{DBLP:journals/tii/ZhangXDZ19}, \cite{yu2018intelligent}.

With the prepared dataset, STGCN is adopted for
training. STGCN is conducted in Python with TensorFlow. As
for the setting of hyperparameters, the batch size is set to 100,
the training epoch is set to 30, the step in one epoch is set
to 48, and the learning rate is set to 0.001. Chebyshev order
K is set to 2. Five spatial-temporal information incorporation
blocks are stacked and the fusion information is utilized for
SVS assessment.

\begin{figure*}[h]
\centering{\includegraphics[height=7.0cm,width=14.5cm]{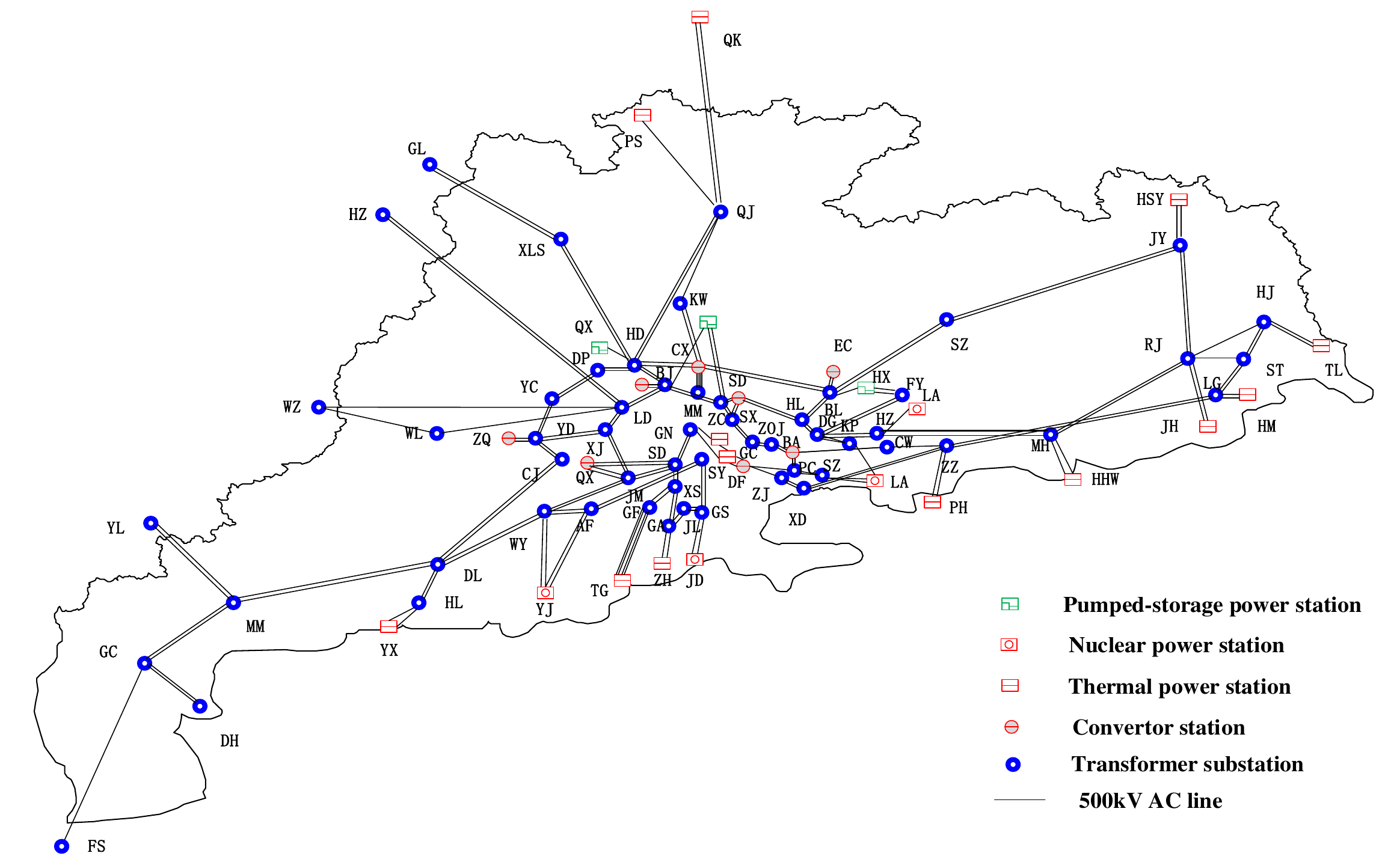}}
\caption{Backbone structure of Guangdong Power grid
\label{fig_sim}}
\end{figure*}

\subsection{STGCN training performances}
This subsection aims to illustrate the effectiveness of the
proposed network and each part in STGCN. First, the effectiveness of the overall network is investigated with training
loss value, training accuracy, and five-fold cross-validation results. Then, according to the data processing order in STGCN,
the effectiveness of each part is illustrated. First, the
locality of Chebyshev filter matrix is illustrated. Then, the effectiveness of spatial-temporal feature
extraction and node layer representation is illustrated by the visualization of the processed data. Finally, this paper verifies the effectiveness of the system layer for
identification of the influences of individual buses on SVS.

\subsubsection{Effectiveness of the overall network}
The distributions of loss value and accuracy in the training process are shown
in Fig. 6. As shown in the two figures, the proposed network shows fast convergence at the beginning of the training
process, and the training accuracy reaches over 95\% after 10
training epochs. With the increase of training time, the loss
value gradually decreases, and the training accuracy increases. The
fluctuation range of loss value and training accuracy also gradually
decreases. Finally, the loss value eventually stables near zero,
and the training accuracy stables near 99\%. Such a training
process illustrates that the training is effective and the model
is developing in the right direction.
With the proposed network, five-fold cross-validation is
conducted. The average training accuracy of the model reaches
99.4\%, and the average testing accuracy reaches 98.8\%. The good performance illustrates the effectiveness of the proposed network.

\begin{figure}[h]
\centering{\includegraphics{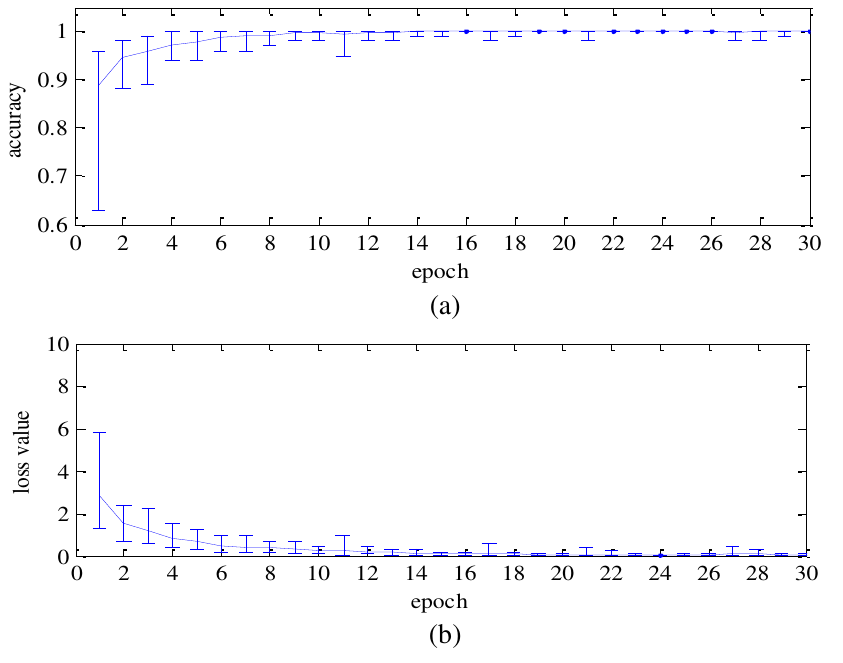}}
\caption{The distributions of loss value and accuracy in training process
\label{fig_sim}}
\end{figure}

\subsubsection{Locality of Chebyshev filter Matrix}
The Chebyshev filter matrix, namely $T_i(\tilde{L})$, is the key element to realize graph convolution. The locality of the Chebyshev filter matrix ensures that the buses only exchange information with their neighboring buses, which is consistent with the spatial characteristics of SVS.

To illustrate the locality of Chebyshev filter matrix, Chebyshev filter matrices of different orders and the corresponding reception fields of MM station
in Guangdong Power Grid are visualized
in Fig. 7, 8, 9. The bus number of MM station is 64,
corresponding to the 64-th row of Chebyshev filter matrices.

As shown in Fig. 7, the position of nonzero elements
in Chebyshev filter matrix at row 64 corresponds to the
stations in the reception field of MM station. It preserves the
locality of the original topology matrix and ensures that the
exchange of information only happens in the bus itself and
its neighboring buses. With the increase of Chebyshev order,
the correspondence between the position of non-zero elements
in Chebyshev filter matrix and stations in the reception field
of one bus remains unchanged, and the reception field grows
bigger, as shown in Fig. 7, 8, 9.

\begin{figure}[h]
\centering{\includegraphics[height=4.2cm,width=8.8cm]{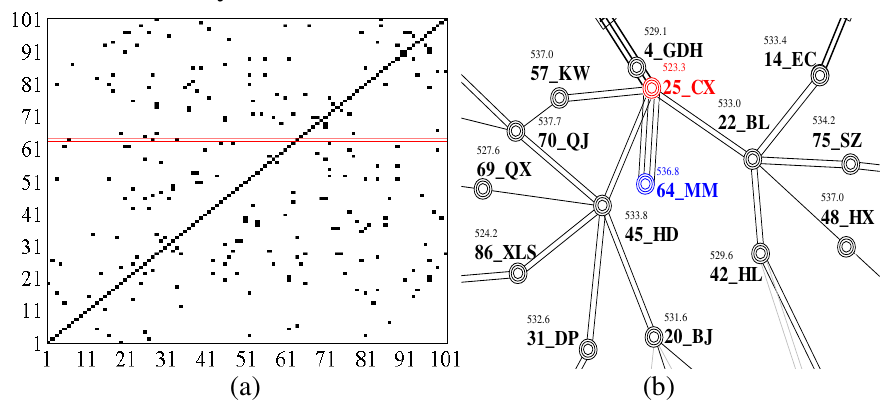}}
\caption{The first-order Chebyshev filter matrix and the reception field of the 64-th station in first-order Chebyshev matrix
\label{fig_sim}}
\end{figure}

\begin{figure}[h]
\centering{\includegraphics[height=4.1cm,width=8.8cm]{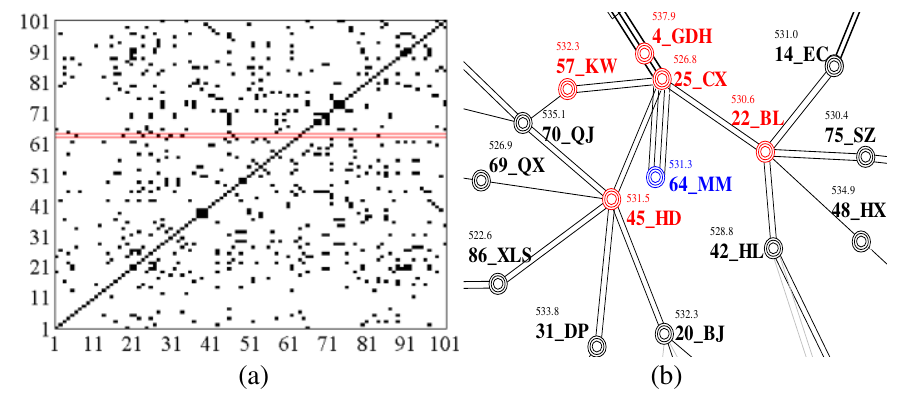}}
\caption{The second Chebyshev filter matrix and the reception field of the 64-th station in second-order Chebyshev matrix
\label{fig_sim}}
\end{figure}

\begin{figure}[h]
\centering{\includegraphics[height=4.1cm,width=8.8cm]{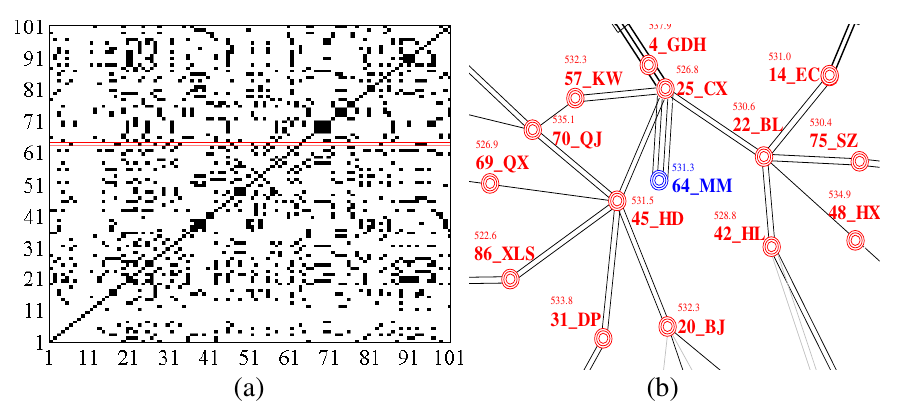}}
\caption{The third-oder Chebyshev filter matrix and the reception field of the 64-th station in third-order Chebyshev matrix
\label{fig_sim}}
\end{figure}

As the reception field of one bus is different according
to Chebyshev order, Chebyshev order $K$ can have a certain
impact on the performances of STGCN. In this paper, $K$ is
set as 2 by multiple tests. Besides, the stacking number of the spatial-temporal information incorporation blocks can also represent the 
reception field of the buses. To capture the spatial-temporal features from different reception fields, the fusion information from the stacked spatial-temporal information incorporation blocks is utilized for the subsequent SVS assessment.

\subsubsection{Spatial-temporal characteristics extraction and node layer representation}
The spatial-temporal characteristics of SVS can be extracted by the spatial-temporal information incorporation blocks. Then, in the node layer block, the transformed multi-channel data can be converted into single-channel representation by applying a weighted summation. Before passing data to the system layer, the data are normalized and the absolute values are taken as their node representations.

To illustrate the effectiveness of spatial-temporal information incorporation blocks and node layer representation block, the resulting representations of 400 cases are visualized in Fig. 10 using t-distributed stochastic neighbor embedding (t-SNE) \cite{dermaaten2008visualizing}, including the data processed by one spatial-temporal information incorporation block $Vidxn_{i,1}$, the data processed by five stacked spatial-temporal information incorporation blocks $Vidxn_{i,5}$, the data processed by spatial-temporal information fusion from the stacked spatial-temporal information incorporation blocks $Vidxz_{i}$, and the data processed by node layer block $Snode_{i}$, $i=1,2,3,...n$.
As shown in Fig. 10, after processed by node layer block, stable cases and unstable cases can be distinguished to a large extent with the transformed dataset. It illustrates the effectiveness of spatial-temporal feature extraction and node layer representation block.

\begin{figure}[h]
\centering{\includegraphics[height=8.2cm,width=8.5cm]{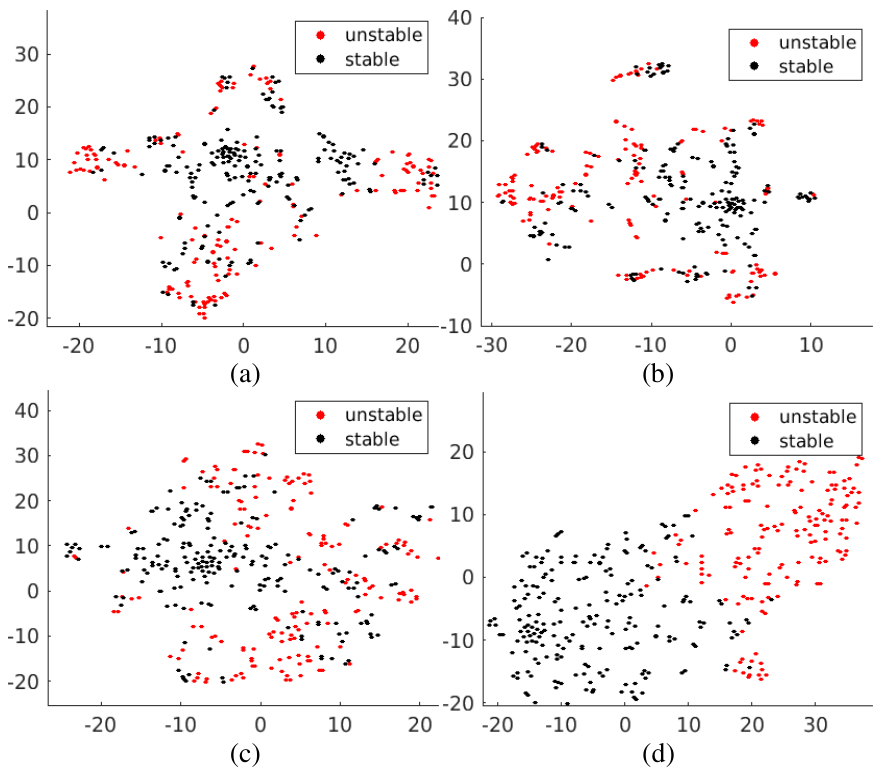}}
\caption{Visualization of the data processed by spatial-temporal information
incorporation blocks and node layer block: (a) Data processed by one
spatial-temporal information incorporation block; (b) Data processed by five
stacked spatial-temporal information incorporation block; (c) Data processed
by spatial-temporal information fusion from the stacked spatial-temporal
information incorporation blocks; (d) Data processed by node layer block
\label{fig_sim}}
\end{figure}

\subsubsection{System layer - identification of the influences of individual
buses on SVS}
After obtaining the node representations of each
case, the simplified differentiable pooling layer is adopted
as the system layer. With the node representation and the
learned parameters $S$ in the system layer, the SVS status in
the target region can be obtained.

The weight values $S$ of the system layer in the well-trained
model can indicate the influences of the corresponding stations
on SVS related to power plants, reactive power injection and
multi-infeed high voltage direct current (HVDC) systems. The
values of the learned weights are shown in Fig. 11. The positive/negative weight value indicates the beneficial/detrimental
influence of the corresponding station on SVS.

Some stations with negative weights present aggregation
behaviors from a spatial perspective, including BA station with
the lowest weight value, XD station with the second-lowest
weight value, and other stations, CW, ZZ, PC, DF. As shown in Fig. 12, the region
corresponding to these stations has multi HVDC systems
infeed, including the HVDC system from
XR station to BA station and the HVDC system from XS
station to DF station. As voltage stability in the multi-HVDC
infeed region is prominent due to the interaction between
direct current (DC) and alternating current (AC) system, it is
consistent with the phenomenon that the stations in that region
present negative weights in the system layer.

The stations with the top three highest weights in the system
layer correspond to DG, GS, and BJ station. For DG and BJ
station, each has a 200 Mvar STATCOM nearby. GS station
has three power plants nearby, which are configured with
excitation and PSS. It is consistent with the general view
that STATCOM and power plants with excitation and PSS
can have positive effects on voltage stability, illustrating the
effectiveness of the parameters learned in the system layer.
\begin{figure}[h]
\centering{\includegraphics[height=6.0cm,width=8.5cm]{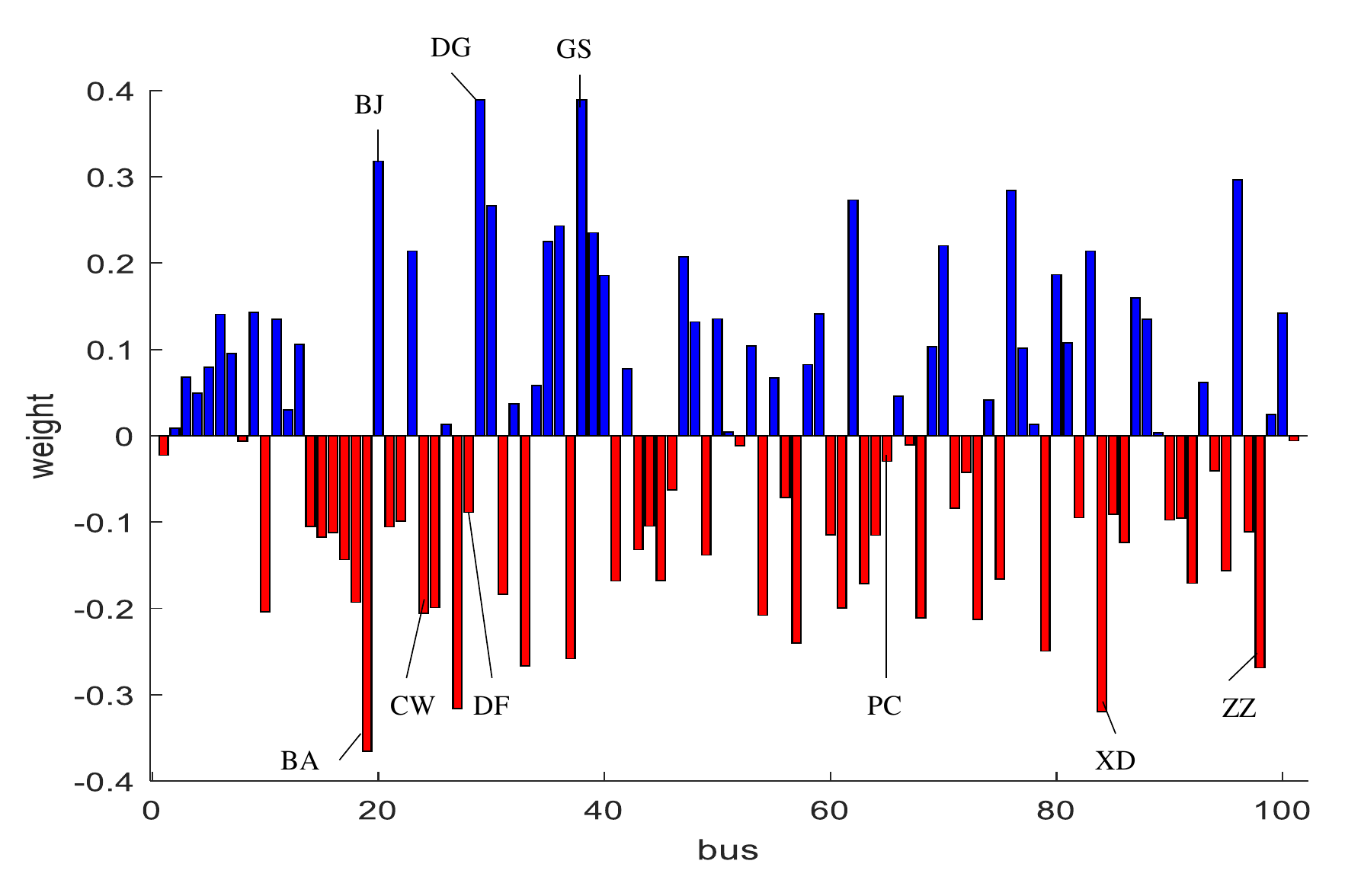}}
\caption{Value of weight in the system layer
\label{fig_sim}}
\end{figure}

\begin{figure}[h]
\centering{\includegraphics[height=3.8cm,width=5.8cm]{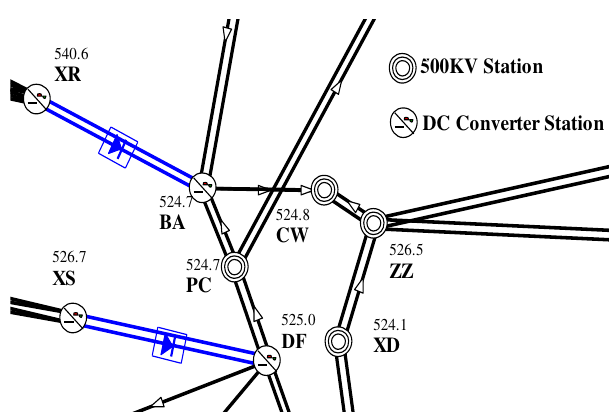}}
\caption{The stations in multi-DC infeed region
\label{fig_sim}}
\end{figure}

\subsection{Performance comparison with other existing stability assessment methods}
In order to further evaluate the performance of the proposed STGCN, it is compared with the existing methods based on post-fault time series, namely the spatial-temporal shapelet
learning method (ST-shapelet) \cite{zhu2020spatialtemporal}, the LSTM-based method
\cite{yu2018intelligent}, and the RVFL-based method \cite{zhang2019a}, under unpredictable faults,
noisy environment, topology changes, and different operating points.

The ST-shapelet method, LSTM-based method and RVFL-based method are the typical methods to deal with post-fault
time series. The ST-shapelet method first generates voltage
animations based on geographical location information and
post-fault time series. Then, it utilizes voltage contour in
the animations to extract comprehensive spatial-temporal time
series. After that, shapelet transform is adopted to convert
the time series dataset into a distance dataset. Finally, it
employs decision tree to conduct classification. The LSTM-based model is composed of an LSTM layer, a hidden dense
neuron layer and a sigmoid function in the end. The LSTM
layer contains 128 cells and can be utilized to handle the
temporal relationship of the dataset. The hidden neuron layer is
used for dimension reduction. After that, the sigmoid function
is adopted to normalize the output. RVFL is a randomized
learning algorithm in the form of single hidden-layer feedforward network. The post-fault time series are flattened and
converted into vectors as the learning input.

\subsubsection{Performance comparison of different approaches for SVS assessment under unpredictable faults}
With the prepared
dataset, the five-fold validation is conducted to compare the
performances of different approaches for SVS assessment, and
the results are shown in Table \uppercase\expandafter{\romannumeral1}.

As shown in Table \uppercase\expandafter{\romannumeral1}, STGCN has the highest training
accuracy and testing accuracy than the other three methods.
The decent performances are attributed to the introduction of
topology with graph convolution and the carefully designed
architecture for SVS assessment. The effectiveness of each part
in the proposed network is verified in the previous subsection.

The performances of LSTM and RVFL-based methods are
lower than the other two methods. This is because these two
methods don’t incorporate any spatial information in their
models in any form, while the ST-shapelet method manages
to incorporate spatial geographic information into the learning model. However, the spatial information represented by
geographic information may not be that suitable compared
to introducing topology into the learning model. Therefore,
the proposed network shows better performances than the ST-shapelet method.
\begin{table}[h]
% increase table row spacing, adjust to taste
\renewcommand{\arraystretch}{1.3}
\caption{Performance comparison of different approaches for SVS assessment under unpredictable faults}
\label{table_example}
\centering
% Some packages, such as MDW tools, offer better commands for making tables
% than the plain LaTeX2e tabular which is used here.
\begin{tabular*}{20pc}{@{\extracolsep{\fill}}lllm{7cm}@{}}
\toprule
\makecell[c]{Model} & \makecell[c]{Training accuracy} & \makecell[c]{Testing accuracy}\\
\midrule
\makecell[c]{STGCN (Proposed)} & \makecell[c]{99.4\%}  & \makecell[c]{98.8\%} \\
\makecell[c]{ST-shapelet}      &\makecell[c]{98.1\%} & \makecell[c]{98.3\%} \\
\makecell[c]{LSTM}             &\makecell[c]{94.8\%} & \makecell[c]{94.4\%} \\
\makecell[c]{RVFL}             & \makecell[c]{97.7\%} & \makecell[c]{97.5\%} \\
\bottomrule 
\end{tabular*}
\end{table}

\subsubsection{Performance comparison of different approaches for voltage stability assessment under noisy environment}
To test the performances of the proposed STGCN under noisy environment, Gaussian
noise is added in the original test dataset. The signal to noise
rate (SNR) is set to 45 dB \cite{brown2016characterizing}. The performances of different
approaches in noisy environments are shown in Table \uppercase\expandafter{\romannumeral2}.

As shown in Table II, STGCN has the highest testing accuracy than the other three methods in noisy environments.
This is due to the effective spatial-temporal feature extraction
and the introduction of the normalization layer and the dropout
layer to the proposed network. As for the ST-shapelet method,
it utilizes key subsequence to distinguish the stability of the
observed system. Therefore, it is relatively less affected by noise
data. The LSTM model doesn’t introduce any module to
address noise, which results in poor performance. The RVFL-based method has the worst performance. It is due to its
shallow structure. RVFL is in the form of single hidden-layer
feedforward network. It cannot extract more effective features
than deep learning and is more susceptible to noise.

When the topology of the observed system contains noise,
it can have a negative impact on the model performance of
STGCN, and the testing accuracy on that occasion is 97.7\%.
It is still relatively higher than the other three methods, which
benefits from the good model design.
\begin{table}[h]
% increase table row spacing, adjust to taste
\renewcommand{\arraystretch}{1.3}
\caption{Performance comparison of different approaches for SVS assessment in noisy environments}
\label{table_example}
\centering
% Some packages, such as MDW tools, offer better commands for making tables
% than the plain LaTeX2e tabular which is used here.
\begin{tabular*}{20pc}{@{\extracolsep{\fill}}llm{7cm}@{}}
\toprule
\makecell[c]{Model}            & \makecell[c]{Testing accuracy}\\
\midrule
\makecell[c]{STGCN (Proposed)} & \makecell[c]{98.6\%}  \\
\makecell[c]{ST-shapelet}      &\makecell[c]{97.4\%}  \\
\makecell[c]{LSTM}             &\makecell[c]{92.0\%}  \\
\makecell[c]{RVFL}             & \makecell[c]{84.0\%}  \\
\bottomrule 
\end{tabular*}
\end{table}

\subsubsection{Performance comparison of different approaches for voltage stability assessment under topology changes}
To test the performance
of the proposed STGCN under topology changes, another
100 cases are generated by setting
different topology changes, fault clearing time, and induction
motor ratios. These cases consider topology changes in two
or three parts. The topology changes are larger than the one-part change in
the previous dataset. The performances of different approaches
under topology changes are shown in Table \uppercase\expandafter{\romannumeral3}.

As shown in Table \uppercase\expandafter{\romannumeral3}, STGCN has the highest testing
accuracy than the other three methods under topology changes.
This is because when the topology of the target region
changes, it will be reflected in the input topology matrix.
Another three methods cannot reveal the changes in topology. The LSTM-based method and RVFL-based method don’t
incorporate any spatial information in their models. The ST-shapelet method only utilizes fixed geographic information
to extract spatial features.

\begin{table}[h]
% increase table row spacing, adjust to taste
\renewcommand{\arraystretch}{1.3}
\caption{Performance comparison of different approaches for SVS assessment under topology changes}
\label{table_example}
\centering
% Some packages, such as MDW tools, offer better commands for making tables
% than the plain LaTeX2e tabular which is used here.
\begin{tabular*}{20pc}{@{\extracolsep{\fill}}llm{7cm}@{}}
\toprule
\makecell[c]{Model}            & \makecell[c]{Testing accuracy}\\
\midrule
\makecell[c]{STGCN (Proposed)} & \makecell[c]{96.0\%}  \\
\makecell[c]{ST-shapelet}      &\makecell[c]{93.0\%}  \\
\makecell[c]{LSTM}             &\makecell[c]{94.0\%}  \\
\makecell[c]{RVFL}             & \makecell[c]{93.0\%}  \\
\bottomrule 
\end{tabular*}
\end{table}

\subsubsection{Performance comparison of different approaches for voltage stability assessment under different operating points}
To test the performance
of the proposed STGCN under different operating points, another
1500 cases are generated by setting
different operating points. Random faults and induction
motor ratios are selected in each case. 1200 cases are utilized to retrain the proposed STGCN, and the remaining cases are for testing. The performances of different approaches
under different operating points are shown in Table \uppercase\expandafter{\romannumeral4}.

As shown in Table \uppercase\expandafter{\romannumeral4}, STGCN has the highest testing
accuracy than the other three methods.
The decent performance is attributed to two aspects. In the input data aspect, the incorporation of network topology information and the temporal data contains full spatial-temporal dynamics in SVS. In the model aspect, the proposed network is designed according to the SVS characteristics. The performance of the ST-shapelet method is higher than the LSTM method and the RVFL method. This is because the ST-shapelet method manages to incorporate spatial information into the learning model with geographic location information.
\begin{table}[h]
% increase table row spacing, adjust to taste
\renewcommand{\arraystretch}{1.3}
\caption{Performance comparison of different approaches for SVS assessment under different operating points}
\label{table_example}
\centering
% Some packages, such as MDW tools, offer better commands for making tables
% than the plain LaTeX2e tabular which is used here.
\begin{tabular*}{20pc}{@{\extracolsep{\fill}}llm{7cm}@{}}
\toprule
\makecell[c]{Model}            & \makecell[c]{Testing accuracy}\\
\midrule
\makecell[c]{STGCN (Proposed)} & \makecell[c]{98.7\%}  \\
\makecell[c]{ST-shapelet}      &\makecell[c]{97.3\%}  \\
\makecell[c]{LSTM}             &\makecell[c]{92.0\%}  \\
\makecell[c]{RVFL}             & \makecell[c]{92.3\%}  \\
\bottomrule 
\end{tabular*}
\end{table}

\section{Conclusion}
Short-term voltage instability phenomenon presents spatial-temporal characteristics, but there is currently no systematic
approach to incorporate such characteristics into the learning
model accurately and effectively. This paper develops STGCN
to incorporate the characteristics of SVS into the learning model.
It employs Chebyshev graph convolution to integrate topology
into the learning model. As Chebyshev filter matrix reveals
locality, it is consistent with the spatial characteristics of SVS. Then, the
one-dimensional convolution is utilized to extract temporal
features. After that, the spatial-temporal features from different
reception fields are extracted by the fusion of spatial-temporal
information incorporation blocks. Finally, the node layer and
system layer are designed to provide the final assessment
result. Test results on Guangdong Power Grid illustrate that compared with the existing typical methods, the
proposed STGCN can achieve higher model accuracy, better
robustness in noisy environments, and better adaptability to
topology changes and different operating points.
Besides, parameters of the system layer in the well-trained
stability assessment model can provide valuable information
about the influences of individual buses on SVS.
% Can use something like this to put references on a page
% by themselves when using endfloat and the captionsoff option.

%\IEEEtriggeratref{8}
%\IEEEtriggercmd{\enlargethispage{-5in}}
%\begin{thebibliography}{1}
\ifCLASSOPTIONcaptionsoff
  \newpage
\fi
\small
\bibliographystyle{IEEEtran}
\bibliography{IEEEabrv,STGCN629}
%\end{thebibliography}
%\begin{thebibliography}{1}

\end{document}